\documentclass[conference]{IEEEtran}
\IEEEoverridecommandlockouts
\usepackage{amsmath,amssymb,amsfonts}
\usepackage{algorithmic}
\usepackage{graphicx}
\usepackage{textcomp}
\usepackage{xcolor}
\def\BibTeX{{\rm B\kern-.05em{\sc i\kern-.025em b}\kern-.08em
    T\kern-.1667em\lower.7ex\hbox{E}\kern-.125emX}}

\usepackage{accents}
\newlength{\dhatheight}
\newcommand{\doublehat}[1]{%
	\settoheight{\dhatheight}{\ensuremath{\hat{#1}}}%
	\addtolength{\dhatheight}{-0.35ex}%
	\hat{\vphantom{\rule{1pt}{\dhatheight}}%
		\smash{\hat{#1}}}}

\usepackage{subfig}
\usepackage{siunitx}
\usepackage{bmpsize}
\usepackage[abs]{overpic}
\usepackage{adjustbox}
\usepackage{booktabs,array}
\usepackage{multirow}
\usepackage{balance}
\renewcommand{\thetable}{\Roman{table}}
\def\tablename{TABLE}

\usepackage{nohyperref}  
\usepackage{url} 

\DeclareRobustCommand*{\IEEEauthorrefmark}[1]{%
	\raisebox{0pt}[0pt][0pt]{\textsuperscript{\footnotesize #1}}%
}

\makeatletter
\renewcommand\footnotesize{%
	\@setfontsize\footnotesize\@ixpt{9.5}%
	\abovedisplayskip 8\p@ \@plus2\p@ \@minus4\p@
	\abovedisplayshortskip \z@ \@plus\p@
	\belowdisplayshortskip 4\p@ \@plus2\p@ \@minus2\p@
	\def\@listi{\leftmargin\leftmargini
		\topsep 4\p@ \@plus2\p@ \@minus2\p@
		\parsep 2\p@ \@plus\p@ \@minus\p@
		\itemsep \parsep}%
	\belowdisplayskip \abovedisplayskip
}
\makeatother

\renewcommand{\baselinestretch}{0.945}
\begin{document}

\title{Unsupervised Medical Image Translation Using Cycle-MedGAN}

	\author{\IEEEauthorblockN{Karim Armanious\IEEEauthorrefmark{1}\textsuperscript{,}\IEEEauthorrefmark{2},
		Chenming Jiang\IEEEauthorrefmark{1},
		Sherif Abdulatif\IEEEauthorrefmark{1},
		Thomas K\"ustner\IEEEauthorrefmark{1}\textsuperscript{,}\IEEEauthorrefmark{2}\textsuperscript{,}\IEEEauthorrefmark{3},
		Sergios Gatidis\IEEEauthorrefmark{2},
		Bin Yang\IEEEauthorrefmark{1}}
	\IEEEauthorblockA{\IEEEauthorrefmark{1}University~of~Stuttgart~,~Institute~of~Signal~Processing~and~System~Theory,~Stuttgart,~Germany\\
		\IEEEauthorrefmark{2}University~of~T\"ubingen,~Department~of~Radiology,~T\"ubingen,~Germany\\
		\IEEEauthorrefmark{3}King's~College~London,~Biomedical~Engineering~Department,~London,~England}}

\maketitle

\begin{abstract}

Image-to-image translation is a new field in computer vision with multiple potential applications in the medical domain. However, for supervised image translation frameworks, co-registered datasets, paired in a pixel-wise sense, are required. This is often difficult to acquire in realistic medical scenarios. On the other hand, unsupervised translation frameworks often result in blurred translated images with unrealistic details. In this work, we propose a new unsupervised translation framework which is titled Cycle-MedGAN. The proposed framework utilizes new non-adversarial cycle losses which direct the framework to minimize the textural and perceptual discrepancies in the translated images. Qualitative and quantitative comparisons against other unsupervised translation approaches demonstrate the performance of the proposed framework for PET-CT translation and MR motion correction.
\end{abstract}


\section{Introduction}
In recent years, the machine learning community has achieved tremendous leaps in performance. This owes to increasingly available computational resources and open-source access to large datasets. From another perspective, radiological scans are vital tools in modern medicine. They enable diagnostics, disease tracking and patient treatment monitoring. This has led to the utilization of recent advances in computer vision, especially Deep Convolutional Neural Networks (DCNNs), in the field of medical image analysis. For example, DCNNs have been adapted for lesion classification in Magnetic Resonance Images (MRI) \cite{4}, 3D image segmentation \cite{6} and anomaly detection \cite{8} among other applications \cite{10,11}.


A branch of deep learning is generative models which are utilized for dataset generation and augmentation. Amongst them, Generative Adversarial Networks (GANs) \cite{12} are the prominent choice, with a large body of research focusing on theoretical and architectural analysis \cite{13,14}. In 2016, the pix2pix framework, a supervised GAN-based framework, has introduced the task of image-to-image translation from a source domain image, e.g. a day-time image, to a corresponding target domain image, e.g. a night-time image, provided that both domains have the same underlying structure \cite{16}. This task has been adapted to the field of medical image analysis by using pix2pix for applications such as low-dose Computed Tomograph (CT) denoising \cite{17}, Positron Emission-computed Tomography (PET) to MR translation \cite{18}, splenomegaly segmentation \cite{19} and MR to CT translation \cite{20}. Other specialized architectures have been introduced for tasks such as compressed sensing MR reconstruction \cite{21} and retinal image super-resolution \cite{22}. Recently, we proposed MedGAN as a new framework for image translation \cite{26}. It extends pix2pix with a cascaded U-net generator architecture \cite{23} and additional non-adversarial losses, such as perceptual \cite{24} and style transfer loss functions \cite{25}. Since then it has been applied to tasks such as PET denoising \cite{26}, MR motion artifacts correction \cite{27} and medical image in-painting \cite{277}.

However, these methods are supervised. In other words, training such models requires paired datasets where the images from the source domain are paired in a pixel-wise sense with the corresponding images in the target domain. Nevertheless, the acquisition of such paired datasets in real-life situations is often challenging. This is due to difficulties in obtaining co-registered cross-modality data from different scanners and acquisition sequences or multi-modal data for some organs such as the heart due to technical challenges or the required extensive planning and acquisition. Consequently, unsupervised image translation techniques, which are trainable with no paired examples but with image samples from both domains, are especially important for medical image translation.

Several methods for unsupervised image-to-image translation have been developed. UNIT is an unsupervised translation framework which maps the input-target images into a common latent space using a pair Variational Auto Encoder-GANs (VAE-GANs) before reconstruction in the desired image domain \cite{29}. It has been utilized in the medical domain for the translation of T1-weighted and T2-weighted MR scans \cite{30}. Cycle-GAN is another unsupervised translation approach which is based on the combination of adversarial losses and the pixel-wise cycle-consistency loss \cite{31}. It has been adapted for medical translation tasks such as CT to MR bidirectional-translation \cite{33} and CT denoising \cite{34}. Other unsupervised frameworks exist with an overview available in \cite{35}.

\begin{figure*}[!ht]
	\centering
	
	\includegraphics[width=0.79\textwidth]{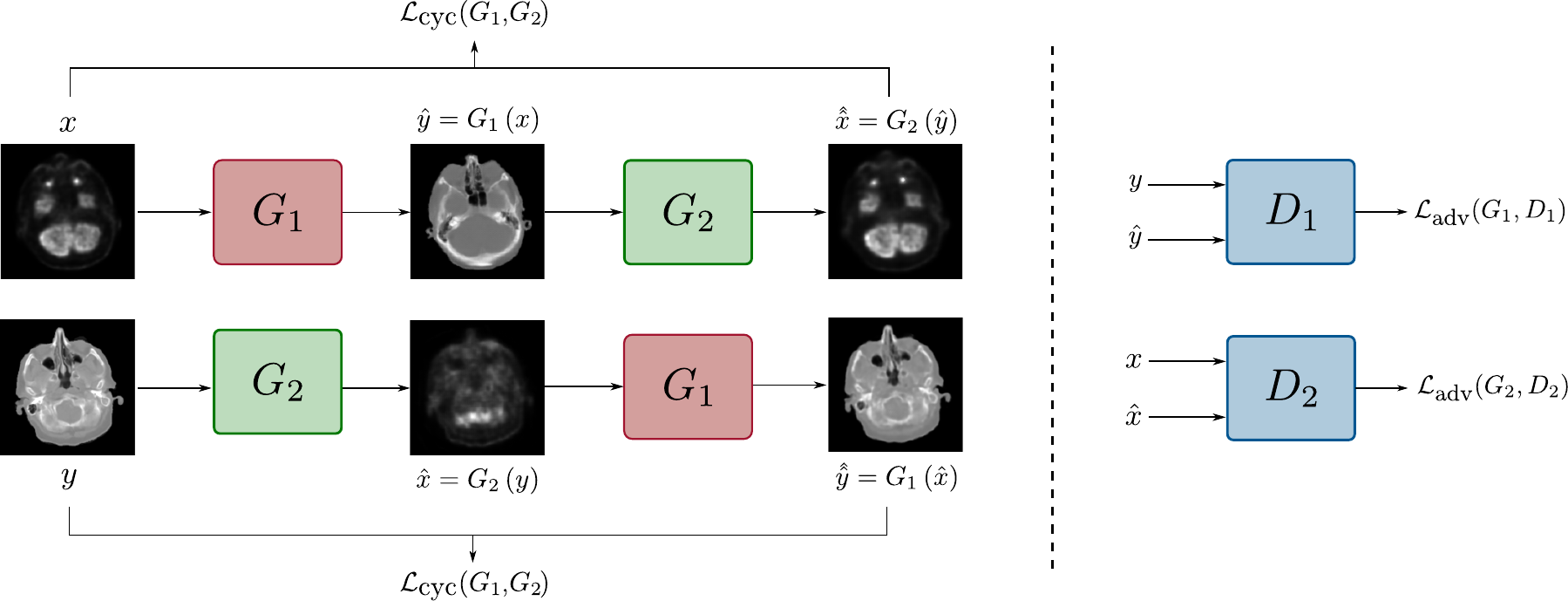}
	
	\caption{An overview of the Cycle-GAN framework for unpaired image translation. $x$ and $y$ are unpaired images randomly sampled from their respective domains.}
	\label{2}
	\vspace{-4mm}
\end{figure*}

In this work, we introduce a new framework for unsupervised medical image translation titled Cycle-MedGAN. This work expands the Cycle-GAN framework by introducing two new non-adversarial loss functions analogous to the perceptual and style transfer losses utilized in MedGAN. However, unlike MedGAN, the calculation of such losses does not require any explicit pairing of the input datasets during training. The training procedure is unsupervised using unpaired data, while validation is conducted on paired datasets.To validate the performance of the proposed framework, qualitative and quantitative comparisons against unsupervised frameworks, such as Cycle-GAN and UNIT, are conducted. The comparisons are carried out on the two medical tasks, MR motion artifact correction and PET to CT translation.  

\section{Method}

The proposed Cycle-MedGAN framework is based on the traditional Cycle-GAN framework with the inclusion of new non-adversarial losses. The baseline Cycle-GAN model is illustrated in Fig. 1.

\subsection{The Cycle-GAN Framework}
Cycle-GAN is an unsupervised framework which allows bidirectional translation between the source domain $X$, e.g. PET images, and the target domain $Y$, e.g. CT images. It consists of two mapping functions $G_1: X\rightarrow{Y}$ and $G_2: Y\rightarrow{X}$, where $G_1$ and $G_2$ are two generator networks parametrized using DCNNs. Each of the generator networks is trained adversarially using a corresponding discriminator network, $D_1$ and $D_2$. For illustration, the first generator network $G_1$ receives as input a source domain image, $x \in X$, and outputs a synthetic translation, $\hat{y} = G_1(x)$. $D_1$ receives as input both the synthetic output $\hat{y}$ and an unpaired image randomly sampled from the desired target domain, $y \in Y$. The two networks, $G_1$ and $D_1$, are pitted against each other in competition, where $D_1$ acts as a binary classifier attempting to distinguish between the translated samples and the target domain samples. On the other hand, $G_1$ attempts to improve the quality of the translated output, thus deceiving the discriminator. This training procedure is formulated as a min-max optimization task over the adversarial loss function $\mathcal{L}_{\small\textrm{adv}}(G_1,D_1)$:
\begin{equation}
\begin{split}
\min_{G_1} \max_{D_1} \mathcal{L}_{\small\textrm{adv}}(G_1,D_1) = & \; \mathbb{E}_{y} \left[\textrm{log} D_1(y) \right] + \\
& \; \mathbb{E}_{x} \left[\textrm{log} \left( 1 - D_1\left(G_1(x)\right) \right) \right]
\end{split}
\end{equation}
and $\mathcal{L}_{\small\textrm{adv}}(G_2,D_2)$ is the analogous loss function for the second pair of networks, $G_2$ and $D_2$, formed by replacing the input images as $y$ and the translated outputs as $\hat{x}$.

 Training the framework merely with the adversarial losses is not sufficient since it may lead to mode collapse, where a set of different input images are mapped into a single image in the target domain \cite{31}. Therefore, an additional constraint regularizing the mapping functions is essential. This is achieved by Cycle-GAN which enforces the two mapping functions, $G_1$ and $G_2$, to be cycle-consistent with each other. In other words the two generator networks should invert each other such that $\doublehat{x} = G_2(G_1(x)) \approx x$ and $\doublehat{y} = G_1(G_2(y)) \approx y$. This behaviour can be incentivized by using the pixel-wise cycle-consistency loss for both generators:
\begin{equation}
\begin{split}
 \mathcal{L}_{\textrm{cyc}}(G_1, G_2)= & \; \mathbb{E}_{x} \left[ \lVert{ x - G_2\left(G_1(x)\right) }\rVert_1 \right]\\
& \; \mathbb{E}_{y} \left[ \lVert{ y - G_1\left(G_2(y)\right) }\rVert_1 \right]
\end{split}
\end{equation}
\subsection{Non-Adversarial Cycle Losses}
\vspace{-0mm}
\begin{figure*}[!ht]
	\centering
	
	\includegraphics[width=0.77\textwidth]{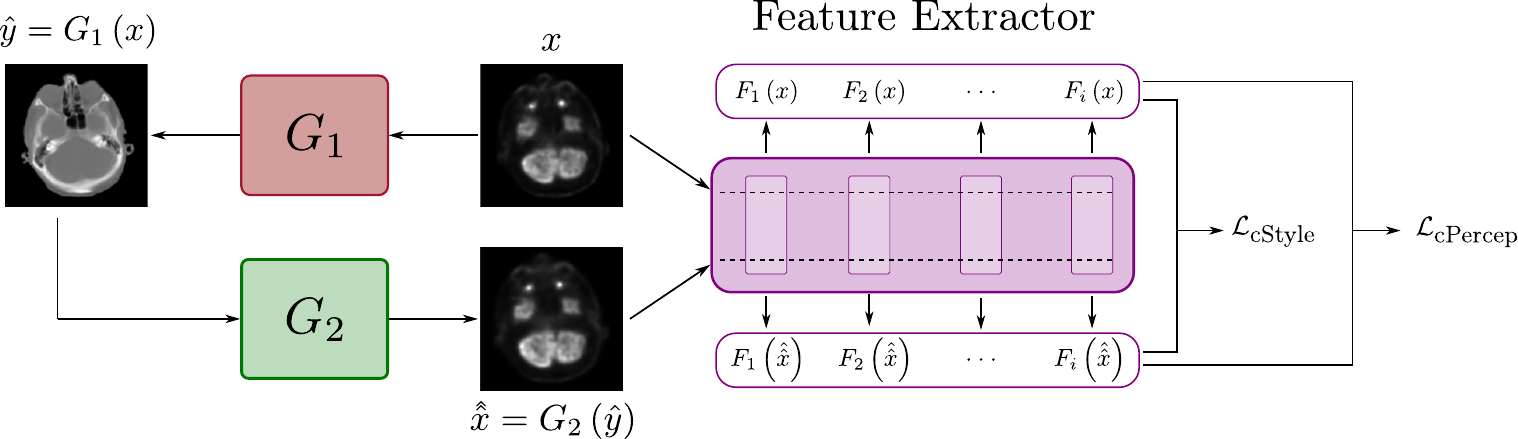}
	
	\caption{An illustration of the proposed cycle non-adversarial loss functions calculated using a pre-trained feature extractor.}
	\label{2}
	\vspace{-4mm}
\end{figure*}
Cycle-GAN relies on the cycle-consistency loss to avoid mismatches which could occur due to unsupervised training using unpaired images. However, it has been discussed in the literature that pixel-wise losses fail to capture the perceptual aspect of human judgement on image quality \cite{36}. Thus, when used in translation tasks, they often lead to results which lack sharpness and fine-detailed structures \cite{24,25}. To circumvent this issue, feature-based loss functions were introduced as additional constraints to enhance the quality of translated output quality. For instance, the MedGAN framework utilized a combination of perceptual and style transfer losses \cite{26}. However, for unsupervised image translation, the utilization of such loss functions is not viable. An unpaired translation model cannot be trained by penalizing the feature-based deviation of the translated image from the unknown ground truth image.

In this work, we propose an adaptation of the above feature-based loss functions for the task of unsupervised image translation. The penalized deviation is instead between the input images, $x$ or $y$, and the cycle-reconstructed images, $\doublehat{x}$ or $\doublehat{y}$. This process is illustrated in Fig. 2. The first proposed loss function is the cycle-perceptual loss, $\mathcal{L}_{\textrm{cPercep}}$. Analogous to the perceptual loss introduced in \cite{26,24}, this loss aims at minimizing the perceptual discrepancies and enhancing the global consistency of the output images. This is achieved by extracting intermediate feature maps, using a pre-trained feature extractor network, for both the input and the cycle-reconstructed images. The cycle-perceptual loss then calculated as the mean absolute error (MAE) between the extracted feature maps for both generators:
 \begin{equation}
\begin{split}
\mathcal{L}_{\small\textrm{cPercep}} = \sum_{i = 0}^{L} \lambda_{cp,i}   \left(\lVert{F_i\left(x\right) - F_i\left(\doublehat{x}\right)}\rVert_1  +  \lVert{F_i\left(y\right) - F_i\left(\doublehat{y}\right)}\rVert_1\right)
\end{split}
\end{equation}\vspace{-0mm}
where $F_i$ and $F_j$ indicate the extracted feature maps from the $i^{\textrm{th}}$ layer of the feature extractor network. $L$ is the total number of layers, $\lambda_{cp,i}$ is the weight given to each layer.

The second proposed loss function is the cycle-style loss, which is typically utilized for style transfer applications \cite{25}. This loss aims at matching the texture, style and fine details of the input images onto the cycle-reconstructed images. As a result, this motivates the generator architectures to produce more detailed translated outputs. The cycle-style loss is computed by first calculating the feature map correlations over the depth dimension. For $G_1$, this is represented by the Gram matrices, $Gr_i(x)$ and $Gr_i(\doublehat{x})$, whose elements are defined as:
 \begin{equation}
Gr_i(x)_{m,n} = \frac{1}{h_i w_i d_i} \sum_{h = 1}^{h_i} \sum_{w = 1}^{w_i} F_{i}(x)_{h,w,m} F_{i}(x)_{h,w,n}
\end{equation}
where $h_i$, $w_i$ and $d_i$ are the spatial height, width and depth of the extracted feature map of the $i^\textrm{th}$ layer of the feature extractor network. 

The style loss is then calculated as the weighted average of the squared Frobenius norm of the Gram matrices:
 \begin{equation}
\begin{split}
\mathcal{L}_{\small\textrm{cStyle}}= & \;  \; \; \, \sum_{i = 1}^{L} \lambda_{cs,i}  \frac{1}{4 d_i^2} \Big( \lVert{Gr_i\left(x\right) - Gr_i\left(\doublehat{x}\right)}\rVert_F^{2}\\
& +   \lVert{Gr_i\left(y\right) - Gr_i\left(\doublehat{y}\right)}\rVert_F^{2} \Big)
\end{split}
\end{equation}
with $\lambda_{cs,i}$ is the weight given to the Gram matrices of the $i^{\textrm{th}}$ layer.

For the Cycle-MedGAN framework, a combination of the adversarial, cycle-consistency, perceptual consistency and style consistency losses is utilized. The final min-max optimization task is given by:
\begin{equation}
	\begin{split}
	\min_{G_1,G_2} \max_{D_1,D_2} \mathcal{L} = & \mathcal{L}_{\small\textrm{adv}}(G_1,D_1) + \mathcal{L}_{\small\textrm{adv}}(G_2,D_2) + \lambda_{cP}\mathcal{L}_{\small\textrm{cPercep}}\\
	& + \lambda_{cyc}\mathcal{L}_{\textrm{cyc}}(G_1,G_2)+ \lambda_{cS}\mathcal{L}_{\small\textrm{cStyle}}
	\end{split}
\end{equation}
where $\lambda_{cP}$, $\lambda_{cyc}$ and $\lambda_{cS}$ are the weights given for the cycle-perceptual, cycle-consistency and cycle-perceptual losses respectively.

\section{Datasets and Experiments}

The Cycle-MedGAN framework was evaluated on two different tasks, PET to CT translation and the correction of motion artifacts in MR. For PET-CT translation, a dataset of the head region from 46 anonymized volunteers was acquired using a joint PET-CT scanner (Siemens Biograph mCT). For training, 1935 two-dimensional slices from 38 patients were utilized while the remaining 420 slices from 8 separate patients were used for validation. For the second application, T1-weighted MR data for the head region was acquired for 17 anonymized volunteers using a 3T MR scanner (Siemens Biograph mMT) with a fast spin echo sequence. Two different scans were acquired for each volunteer, one with voluntary rigid motion (head tilting) of the head and another under resting conditions \cite{37}. Another 980 slices from 14 patients were used for training and 105 slices from the remaining 3 patients were used for validation. For both applications, the resolution of extracted data slices was re-sampled from their original resolutions to an isotropic voxel size of $1\textrm{mm}^{3}$ and two-dimensional images of pixel dimensions $256 \times 256$ were extracted. 

Analogous to Cycle-GAN, random shuffling was applied between the different subjects in the collected datasets to ensure no explicit pairing between the source and target domains occur during the training procedure. The discriminator from a BiGAN network was utilized as the feature extractor network in all experimentations \cite{38}. The BiGAN was pre-trained on a separate dataset of whole-body CT data for image reconstruction. This was conducted to extract plausible feature maps which would improve the quality of translated images. 

\begin{table}[!t]
	\caption{Quantitative comparison of unsupervised translation techniques\label{t1}}
	\centering
	\setlength\arrayrulewidth{0.05pt}
	\tiny
	\bgroup
	\def\arraystretch{1.15}
	\resizebox{\columnwidth}{!}{%
		\begin{tabular}{r|cccc}
			\hline\hline
			\multirow{2}{*}{Model} & \multicolumn{4}{c}{(a) PET-CT translation}\\ & SSIM & PSNR(dB) & VIF & LPIPS\\
			\hline
			UNIT & 0.8485 & 20.14 & 0.2057 & 0.6762\\
			Cycle-GAN & 0.8963 & 23.35 & 0.3831 & 0.2561
			\\
			Cycle-MedGAN & \textbf{0.9115} & \textbf{24.08} & \textbf{0.4275} & \textbf{0.2233}
			\\
			\hline \hline
			\multirow{2}{*}{Model} & \multicolumn{4}{c}{(b) MR motion correction}\\ & SSIM & PSNR(dB) & MSE & UQI\\
			\hline
			UNIT & 0.6914 & 18.64 & 0.1239 & 0.6953\\
			Cycle-GAN & 0.8011 & 22.39 & 0.3432 & 0.3282
			\\
			Cycle-MedGAN & \textbf{0.8118} & \textbf{22.96} & \textbf{0.3513} & \textbf{0.3029}
			\\
			\hline
		\end{tabular}
	}
	\egroup
	\vspace{-5mm}
\end{table}

\begin{figure*}[t]
	\centering
	\begin{minipage}[t]{0.8\linewidth}
		\centering
		\vspace{7mm}
		\begin{minipage}[t]{0.2\linewidth}
			\centering
			\begin{overpic}[width=0.90\textwidth]%
				{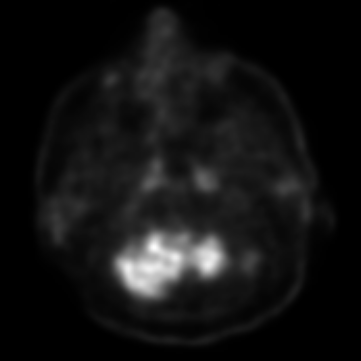}
				\centering
				\put(27,87){Input}
			\end{overpic}	
		\end{minipage}%
		\begin{minipage}[t]{0.6\linewidth}
			\centering
			\begin{overpic}[width=0.30\textwidth]%
				{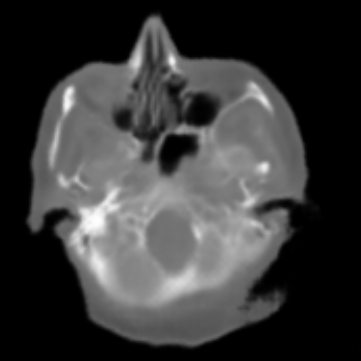}
				\centering
				\put(27,87){UNIT}
			\end{overpic}
			\begin{overpic}[width=0.30\textwidth]%
				{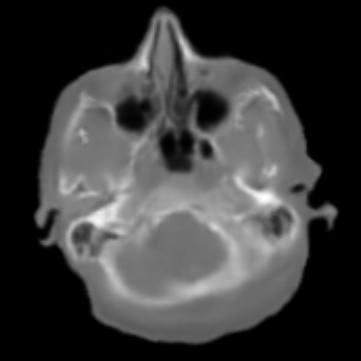}
				\centering
				\put(13,87){Cycle-GAN}
			\end{overpic}
			\begin{overpic}[width=0.30\textwidth]%
				{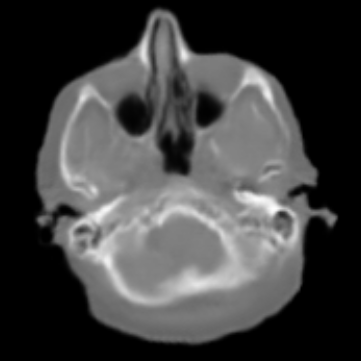}
				\centering
				\put(5,87){Cycle-MedGAN}
			\end{overpic}
		\end{minipage}
		\begin{minipage}[t]{0.182\linewidth}
			\centering
			\begin{overpic}[width=0.99\textwidth]%
				{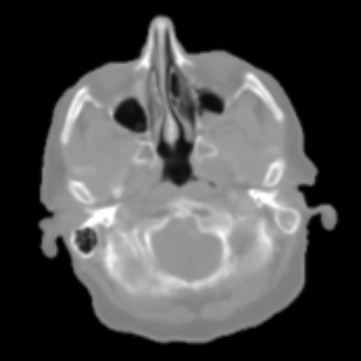}
				\centering
				\put(25,87){Target}
			\end{overpic}		
		\end{minipage}\\
		\vspace{2mm} 
		\begin{minipage}[t]{0.2\linewidth}
			\centering
			\begin{overpic}[width=0.90\textwidth]%
				{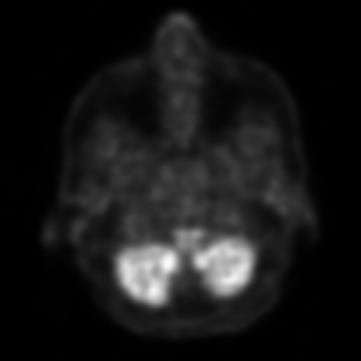}
			\end{overpic}	
		\end{minipage}%
		\begin{minipage}[t]{0.6\linewidth}
			\centering
			\begin{overpic}[width=0.30\textwidth]%
				{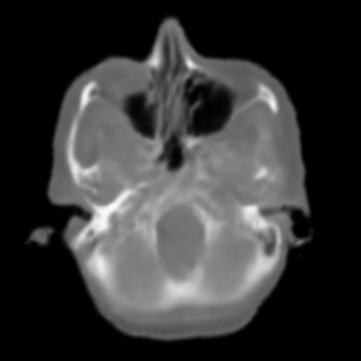}
			\end{overpic}
			\begin{overpic}[width=0.30\textwidth]%
				{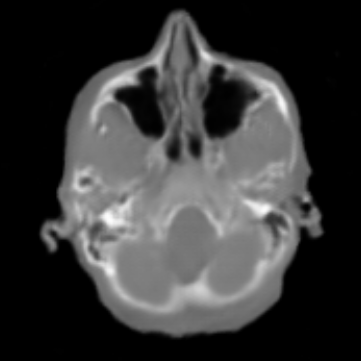}
			\end{overpic}
			\begin{overpic}[width=0.30\textwidth]%
				{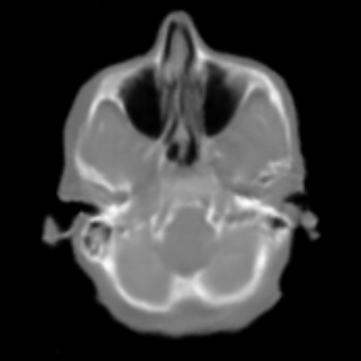}
			\end{overpic}
		\end{minipage}
		\begin{minipage}[t]{0.182\linewidth}
			\centering
			\begin{overpic}[width=0.99\textwidth]%
				{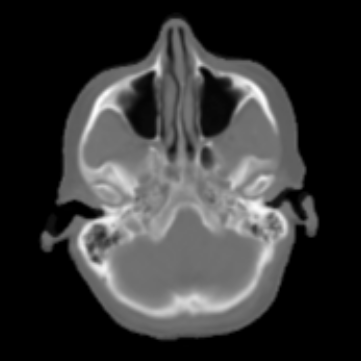}
			\end{overpic}		
		\end{minipage}\\
		\vspace{2mm} 
		\begin{minipage}[t]{0.2\linewidth}
			\centering
			\begin{overpic}[width=0.90\textwidth]%
				{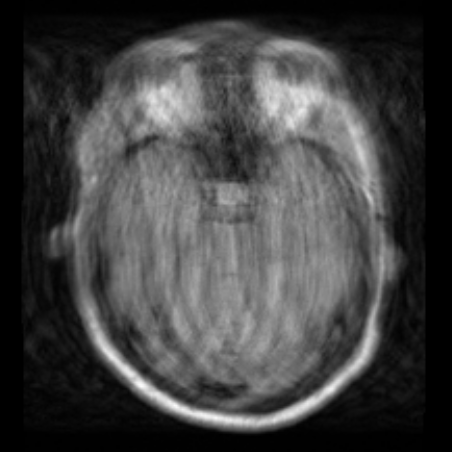}
			\end{overpic}	
		\end{minipage}%
		\begin{minipage}[t]{0.6\linewidth}
			\centering
			\begin{overpic}[width=0.30\textwidth]%
				{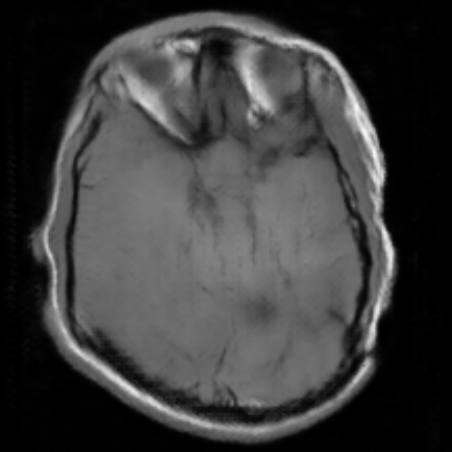}
			\end{overpic}
			\begin{overpic}[width=0.30\textwidth]%
				{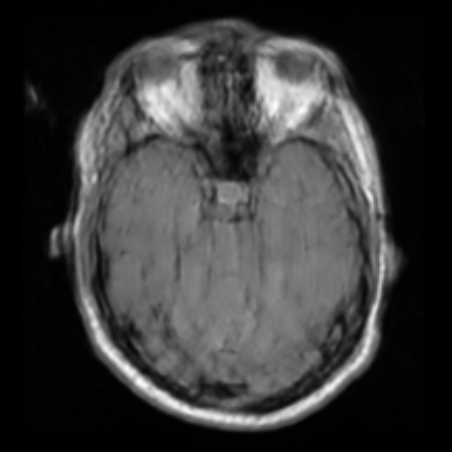}
			\end{overpic}
			\begin{overpic}[width=0.30\textwidth]%
				{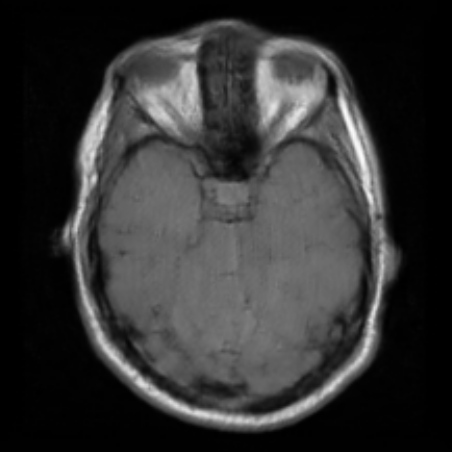}
			\end{overpic}
		\end{minipage}
		\begin{minipage}[t]{0.182\linewidth}
			\centering
			\begin{overpic}[width=0.99\textwidth]%
				{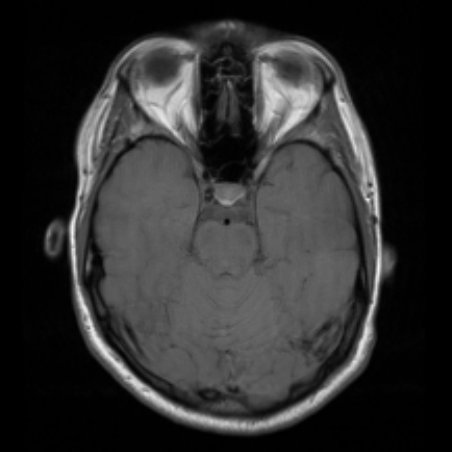}
			\end{overpic}		
		\end{minipage}\\
		\vspace{2mm} 
		\begin{minipage}[t]{0.2\linewidth}
			\centering
			\begin{overpic}[width=0.90\textwidth]%
				{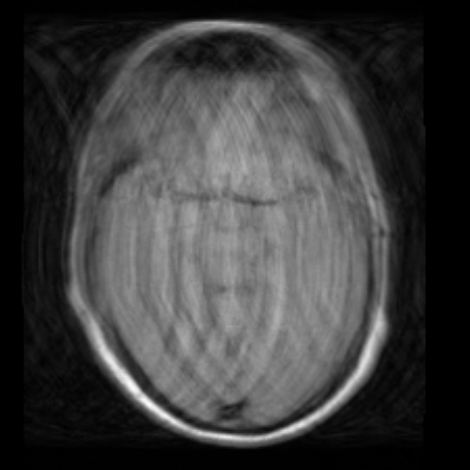}
			\end{overpic}	
		\end{minipage}%
		\begin{minipage}[t]{0.6\linewidth}
			\centering
			\begin{overpic}[width=0.30\textwidth]%
				{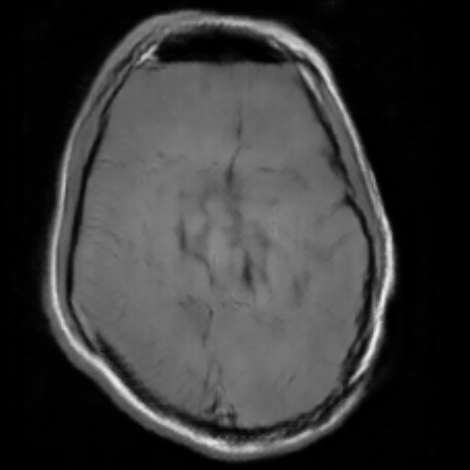}
			\end{overpic}
			\begin{overpic}[width=0.30\textwidth]%
				{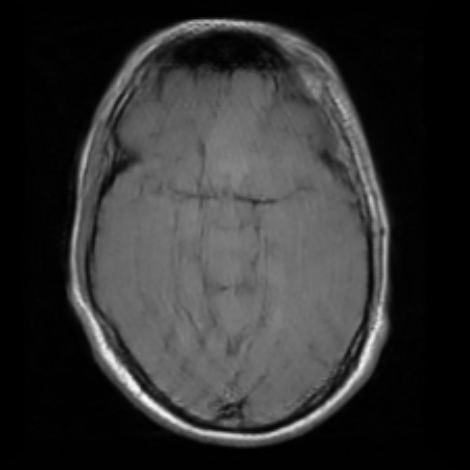}
			\end{overpic}
			\begin{overpic}[width=0.30\textwidth]%
				{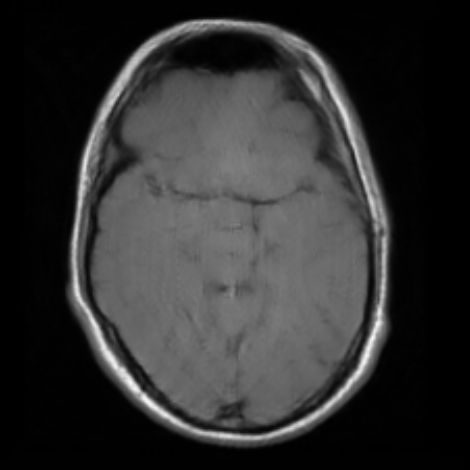}
			\end{overpic}
		\end{minipage}
		\begin{minipage}[t]{0.182\linewidth}
			\centering
			\begin{overpic}[width=0.99\textwidth]%
				{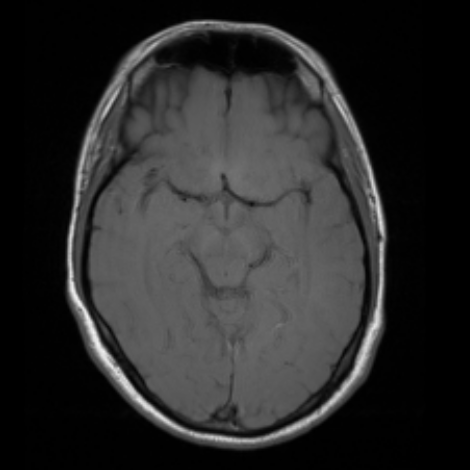}
			\end{overpic}		
		\end{minipage}\\
	\end{minipage}
	\caption{Qualitative comparisons between the Cyclc-MedGAN framework and other unsupervised image translation techniques. The first two rows depict the task of PET to CT translation and last two rows illustrate the correction of MR motion artifacts.}
	\label{5}
	\vspace{-4mm}
\end{figure*}

To evaluate the performance of the proposed framework, qualitative and quantitative comparisons with other unsupervised translation techniques were carried out. More specifically, the UNIT framework \cite{29} and the Cycle-GAN \cite{31} were considered as performance baselines for the comparison. To ensure faithful representation of the baseline methods, verified open-source implementations were utilized along with the recommended hyper-parameters in the original publications \cite{39,40}. Besides the pre-trained feature extractor, the Cycle-MedGAN framework has an identical architecture as the Cycle-GAN framework which is described in details in \cite{31}. The quantitative comparisons utilized the following metrics: Peak Signal to Noise Ratio (PSNR), Structural Similarity Index (SSIM) \cite{41}, Learned Perceptual Image Patch Similarity (LPIPS) \cite{42} and Visual Information Fidelity (VIF) \cite{43}. All models were trained using the ADAM optimizer and a batch size of 64. Training was for 50 epochs, lasting approximately 24 hours, using an NVIDIA Titan X GPU. 

\section{Results and Discussion}\label{AA}

The results of the proposed Cycle-MedGAN framework are presented in Table~\ref{t1} and Fig.~\ref{5} in comparison with UNIT and Cycle-GAN. Qualitatively, the worse performance is exhibited by the UNIT framework. For both medical datasets in the comparative study, UNIT results in inhomogeneous global deformations as well as substantial blurs and distortion in the translated images. This is also reflected quantitatively, with UNIT resulting in the worst scores in Table~\ref{t1} across the chosen metrics. In contrast, the resultant images produced by Cycle-GAN framework have a global structure which closely matches that of the target ground-truth images. However, finer details are not accurately translated by Cycle-GAN, such as the bone structures in the resultant CT images, and the motion blurring due to rigid motion in MR is not completely removed. The proposed Cycle-MedGAN framework builds upon the traditional Cycle-GAN architecture by introducing addition non-adversarial cycle losses to regulate the generator architectures. This results in an enhanced visual quality in the translated images (e.g. sharp edge delineation in MR images). The resultant CT images have noticeably sharper and more consistent bone structures compared to the other frameworks. Additionally, motion blurring due to rigid motion in MR is minimized. This enhancement in performance is analogously reflected quantitatively in Table~\ref{t1} with Cycle-MedGAN surpassing the comparison baselines on all chosen metrics.

Despite the added level of quality by the Cycle-MedGAN framework, the proposed technique is not free from drawbacks. First, the resultant images by the Cycle-MedGAN framework potentially overlook important diagnostic information in the translation process. Thus, the framework is not intended for diagnostic applications but rather for post-processing tasks. An example of such tasks is using the synthetic CT images for PET attenuation correction or the calculation of organ volumes from motion corrupted MR images. Additionally, phase information contains vital information for motion correction in MR. Therefore, in future studies, we plan on expanding the framework with multi-channel three-dimensional inputs for complex-valued data. Furthermore, we plan to expand the comparison study to include performance on clinical applications and the effect of each individual non-adversarial loss in comparison to different loss combinations.


\section{Conclusion}
Cycle-MedGAN is a new framework for unsupervised image translation. It builds upon the widely utilized Cycle-GAN framework with the additional utilization of new non-adversarial cycle loss functions, namely the cycle-perceptual loss and the cycle-style loss. The new loss functions use intermediate feature maps, extracted from a pre-trained feature extractor network, to direct the generator architectures to minimize perceptual and textural discrepancies in the results. Quantitative and qualitative comparisons with other unsupervised translation techniques indicate that the proposed framework enhances the translated outputs for the tasks of PET to CT translation and MR motion correction.

In the future, we plan to enhance the framework from the architectural aspect by incorporating three-dimensional complex-valued data. Additionally, the diagnostic performance of the framework will be investigated by experienced radiologists conducting subjective evaluations of the results.
\bibliographystyle{IEEEtran}


\begin{thebibliography}{}
\providecommand{\url}[1]{#1}
\csname url@samestyle\endcsname
\providecommand{\newblock}{\relax}
\providecommand{\bibinfo}[2]{#2}
\providecommand{\BIBentrySTDinterwordspacing}{\spaceskip=0pt\relax}
\providecommand{\BIBentryALTinterwordstretchfactor}{4}
\providecommand{\BIBentryALTinterwordspacing}{\spaceskip=\fontdimen2\font plus
\BIBentryALTinterwordstretchfactor\fontdimen3\font minus
  \fontdimen4\font\relax}
\providecommand{\BIBforeignlanguage}[2]{{%
\expandafter\ifx\csname l@#1\endcsname\relax
\typeout{** WARNING: IEEEtran.bst: No hyphenation pattern has been}%
\typeout{** loaded for the language `#1'. Using the pattern for}%
\typeout{** the default language instead.}%
\else
\language=\csname l@#1\endcsname
\fi
#2}}
\providecommand{\BIBdecl}{\relax}
\BIBdecl

\end{thebibliography}


\begin{thebibliography}{10}
	\providecommand{\url}[1]{#1}
	\csname url@samestyle\endcsname
	\providecommand{\newblock}{\relax}
	\providecommand{\bibinfo}[2]{#2}
	\providecommand{\BIBentrySTDinterwordspacing}{\spaceskip=0pt\relax}
	\providecommand{\BIBentryALTinterwordstretchfactor}{4}
	\providecommand{\BIBentryALTinterwordspacing}{\spaceskip=\fontdimen2\font plus
		\BIBentryALTinterwordstretchfactor\fontdimen3\font minus
		\fontdimen4\font\relax}
	\providecommand{\BIBforeignlanguage}[2]{{%
			\expandafter\ifx\csname l@#1\endcsname\relax
			\typeout{** WARNING: IEEEtran.bst: No hyphenation pattern has been}%
			\typeout{** loaded for the language `#1'. Using the pattern for}%
			\typeout{** the default language instead.}%
			\else
			\language=\csname l@#1\endcsname
			\fi
			#2}}
	\providecommand{\BIBdecl}{\relax}
	\BIBdecl
	
	\bibitem{4}
	Q.~{Dou} \emph{et~al.}, ``{Multilevel Contextual 3-D CNNs for False Positive
		Reduction in Pulmonary Nodule Detection},'' \emph{IEEE Transactions on
		Biomedical Engineering}, vol.~64, no.~7, pp. 1558--1567, July 2017.
	
	\bibitem{6}
	K.~Kamnitsas \emph{et~al.}, ``{Efficient multi-scale 3D CNN with fully
		connected CRF for accurate brain lesion segmentation},'' \emph{Medical Image
		Analysis}, vol.~36, pp. 61--78, 2017.
	
	\bibitem{8}
	T.~Schlegl \emph{et~al.}, ``{Unsupervised Anomaly Detection with Generative
		Adversarial Networks to Guide Marker Discovery},'' in \emph{Information
		Processing in Medical Imaging (IPMI)}, 2017, pp. 146--157.
	
	\bibitem{10}
	H.~R. Roth \emph{et~al.}, ``{A New 2.5D Representation for Lymph Node Detection
		Using Random Sets of Deep Convolutional Neural Network Observations},'' in
	\emph{Medical Image Computing and Computer-Assisted Interventio (MICCAI)},
	2014, pp. 520--527.
	
	\bibitem{11}
	H.~{Greenspan}, B.~{van Ginneken}, and R.~M. {Summers}, ``{Guest Editorial Deep
		Learning in Medical Imaging: Overview and Future Promise of an Exciting New
		Technique},'' \emph{IEEE Transactions on Medical Imaging}, vol.~35, no.~5,
	pp. 1153--1159, May 2016.
	
	\bibitem{12}
	I.~J. Goodfellow \emph{et~al.}, ``{Generative Adversarial Networks},'' in
	\emph{Conference on Neural Information Processing Systems (NIPS)}, 2014, pp.
	2672--2680.
	
	\bibitem{13}
	T.~Salimans \emph{et~al.}, ``{Improved Techniques for Training GANs},'' in
	\emph{Conference on Neural Information Processing Systems (NIPS)}, 2016, pp.
	2234--2242.
	
	\bibitem{14}
	I.~Gulrajani \emph{et~al.}, ``{Improved Training of Wasserstein GANs},'' in
	\emph{Conference on Neural Information Processing Systems (NIPS)}, 2017, pp.
	5769--5779.
	
	\bibitem{16}
	P.~Isola \emph{et~al.}, ``{Image-to-Image Translation with Conditional
		Adversarial Networks},'' in \emph{Conference on Computer Vision and Pattern
		Recognition (CVPR)}, 2016, pp. 5967--5976.
	
	\bibitem{17}
	J.~M. {Wolterink} \emph{et~al.}, ``{Generative Adversarial Networks for Noise
		Reduction in Low-Dose CT},'' \emph{IEEE Transactions on Medical Imaging},
	vol.~36, no.~12, pp. 2536--2545, Dec 2017.
	
	\bibitem{18}
	H.~Choi and D.~S. Lee, ``{Generation of Structural {MR} Images from Amyloid
		{PET}: Application to {MR}-Less Quantification},'' \emph{Journal of Nuclear
		Medicine}, vol.~59, pp. 1111--1117, 2018.
	
	\bibitem{19}
	Y.~Huo \emph{et~al.}, ``{Splenomegaly segmentation using global convolutional
		kernels and conditional generative adversarial networks},'' in \emph{Medical
		Imaging 2018: Image Processing}.
	
	\bibitem{20}
	H.~E. M. \emph{et~al.}, ``{Generating synthetic CTs from magnetic resonance
		images using generative adversarial networks},'' \emph{Medical Physics},
	vol.~45, no.~8, pp. 3627--3636, 2018.
	
	\bibitem{21}
	T.~M. {Quan}, T.~{Nguyen-Duc}, and W.~{Jeong}, ``{Compressed Sensing MRI
		Reconstruction Using a Generative Adversarial Network With a Cyclic Loss},''
	\emph{IEEE Transactions on Medical Imaging}, vol.~37, no.~6, pp. 1488--1497,
	June 2018.
	
	\bibitem{22}
	D.~Mahapatra \emph{et~al.}, ``{Image Super Resolution Using Generative
		Adversarial Networks and Local Saliency Maps for Retinal Image Analysis},''
	in \emph{Medical Image Computing and Computer-Assisted Intervention
		(MICCAI)}, 2017, pp. 382--390.
	
	\bibitem{26}
	K.~Armanious \emph{et~al.}, ``{{MedGAN}: Medical Image Translation using
		{GANs}},'' \url{http://arxiv.org/abs/1806.06397v1}, 2018, arXiv preprint.
	
	\bibitem{23}
	S.~Shah \emph{et~al.}, ``{Stacked {U-Nets:} A No-Frills Approach to Natural
		Image Segmentation},'' \url{https://arxiv.org/abs/1804.10343}, 2018, arXiv
	preprint.
	
	\bibitem{24}
	C.~Wang \emph{et~al.}, ``{Perceptual Adversarial Networks for Image-to-Image
		Transformation},'' \emph{IEEE Transactions on Image Processing}, vol.~27,
	2018.
	
	\bibitem{25}
	J.Johnson, A.~Alahi, and F.~Li, ``{Perceptual Losses for Real-Time Style
		Transfer and Super-Resolution},'' 2016, pp. 694--711.
	
	\bibitem{27}
	K.~Armanious \emph{et~al.}, ``{Retrospective correction of Rigid and Non-Rigid
		MR motion artifacts using GANs},'' \url{https://arxiv.org/abs/1809.06276},
	2019, accepted to IEEE International Symposium for Biomedical Images (ISBI).
	
	\bibitem{277}
	K.~Armanious, Y.~Mecky, S.~Gatidis, and B.~Yang, ``{Adversarial Inpainting of
		Medical Image Modalities},'' \url{https://arxiv.org/abs/1810.06621}, 2019,
	accepted to IEEE International Conference on Acoustics, Speech, and Signal
	Processing (ICASSP).
	
	\bibitem{29}
	M.~Liu, T.~Breuel, and J.~Kautz, ``{Unsupervised Image-to-Image Translation
		Networks},'' in \emph{Conference on Neural Information Processing Systems
		(NIPS)}, 2017, pp. 700--708.
	
	\bibitem{30}
	P.~Welander, S.~Karlsson, and A.~Eklund, ``{Generative Adversarial Networks for
		Image-to-Image Translation on Multi-Contrast MR Images - A Comparison of
		CycleGAN and UNIT},'' \url{https://arxiv.org/abs/1806.07777}, 2018, arXiv
	preprint.
	
	\bibitem{31}
	J.~{Zhu} \emph{et~al.}, ``{Unpaired Image-to-Image Translation Using
		Cycle-Consistent Adversarial Networks},'' in \emph{IEEE International
		Conference on Computer Vision (ICCV)}, Oct 2017, pp. 2242--2251.
	
	\bibitem{33}
	C.~Jin \emph{et~al.}, ``{Deep CT to MR Synthesis using Paired and Unpaired
		Data},'' \url{https://arxiv.org/abs/1805.10790}, 2018, arXiv preprint.
	
	\bibitem{34}
	E.~Kang \emph{et~al.}, ``{Cycle-consistent adversarial denoising network for
		multiphase coronary CT angiography},'' \emph{Medical Physics}, vol.~46,
	no.~2, pp. 550--562, 2019.
	
	\bibitem{35}
	X.~Yi, E.~Walia, and P.~Babyn, ``{Generative Adversarial Network in Medical
		Imaging: A Review},'' \url{https://arxiv.org/abs/1809.07294}, 2018, arXiv
	preprint.
	
	\bibitem{36}
	D.~Pathak \emph{et~al.}, ``{Context Encoders: Feature Learning by
		Inpainting},'' \emph{IEEE Conference on Computer Vision and Pattern
		Recognition (CVPR)}, pp. 2536--2544, 2016.
	
	\bibitem{37}
	T.~K{\"u}stner \emph{et~al.}, ``Automated reference-free detection of motion
	artifacts in magnetic resonance images,'' in \emph{Magnetic Resonance
		Materials in Physics, Biology and Medicine}, vol.~31, 2018, pp. 243--256.
	
	\bibitem{38}
	J.~Donahue, P.~Krahenb{\"u}hl, and T.~Darrell, ``Adversarial feature
	learning,'' in \emph{International Conference on Learning Representations
		(ICLR)}, 2017.
	
	\bibitem{39}
	J.~Kim, ``{UNIT} implementation,''
	\url{https://github.com/taki0112/UNIT-Tensorflow}.
	
	\bibitem{40}
	X.~Hu, ``{Cycle-GAN} implementation,''
	\url{https://github.com/xhujoy/CycleGAN-tensorflow}.
	
	\bibitem{41}
	Z.~Wang \emph{et~al.}, ``{Image quality assessment: from error visibility to
		structural similarity},'' in \emph{IEEE Transactions on Image Processing},
	vol.~13, 2004, pp. 600--612.
	
	\bibitem{42}
	R.~Zhang \emph{et~al.}, ``{The Unreasonable Effectiveness of Deep Features as a
		Perceptual Metric},'' in \emph{Conference on Computer Vision and Pattern
		Recognition}, 2018.
	
	\bibitem{43}
	H.~R. Sheikh and A.~C. Bovik, ``{Image information and visual quality},'' in
	\emph{IEEE Transactions on Image Processing}, vol.~15, 2006, pp. 430--444.
	
\end{thebibliography}
\end{document}